\documentclass{article}

\usepackage[dblblindworkshop, final]{neurips_2025}
\usepackage{amsmath}

\usepackage{amsmath}
\usepackage{amssymb}
\usepackage{wrapfig} % Already included, used for wrapping
\usepackage{mathtools}
\usepackage{amsthm}
\usepackage{wrapfig}    % wrap figures
\usepackage{tcolorbox}    % wrap figures
\newcommand{\op}[1]{\mathrm{#1}}
\newcommand{\SubExpr}{\operatorname{SubExpr}}
\newcommand{\win}[1]{\textcolor{red}{#1}}

\workshoptitle{MATH-AI}

\usepackage[utf8]{inputenc} % allow utf-8 input
\usepackage[T1]{fontenc}    % use 8-bit T1 fonts
\usepackage{hyperref}       % hyperlinks
\usepackage{url}            % simple URL typesetting
\usepackage{booktabs}       % professional-quality tables
\usepackage{amsfonts}       % blackboard math symbols
\usepackage{nicefrac}       % compact symbols for 1/2, etc.
\usepackage{microtype}      % microtypography
\usepackage{xcolor}         % colors

\title{Curiosity-driven RL for symbolic equation solving}

\author{%
  Kevin P.\ O'Keeffe\\
  Starling Research Institute\\
  Seattle, WA, USA\\
  \texttt{kevin.p.okeeffe@gmail.com}
}

\begin{document}

\maketitle

\begin{abstract}
We explore if RL can be useful for symbolic mathematics. Previous work showed contrastive learning can solve linear equations in one variable. We show model-free PPO \cite{schulman2017proximal} augmented with curiosity-based exploration and graph-based actions can solve nonlinear equations such as those involving radicals, exponentials, and trig functions. Our work suggests curiosity-based exploration may be useful for general symbolic reasoning tasks. 
\end{abstract}

%%%%%%%%%%%%%%%%%%%%%%%%%%%%%%%%%%%%%%%%%%%%%%%%%%%%%%%%%%%%%%%%
\section{Introduction}
Reinforcement learning (RL) has been applied to diverse fields \citep{ng2006autonomous, degrave2022magnetic, vinyals2019grandmaster} but has not yet been widely adopted in symbolic mathematics, where agents perform tasks like solving algebraic equations or evaluating integrals analytically. Traditional approaches to symbolic mathematics rely on hand-engineered systems like Mathematica or Maple. An RL-based approach, where agents learn transformations autonomously, could reduce manual curation and find novel solution techniques.

Applying RL to symbolic math is difficult because the state space is combinatorially large, as each equation can branch into numerous sub-expressions. The action space is also large and dynamic: at every step, the agent can apply various algebraic manipulations to different sub-expressions, so the number and type of actions change with the equation's form.

We here show RL agents combined with curiosity-based exploration and a graph-based action space can overcome these challenges. The agents learn to solve a wide range of algebraic equations such as those involving radicals and trig functions. Though perhaps simple from a human perspective, solving such equations using RL is non-trivial and to our knowledge has never before been done. Moreover, they are a key building block for more advanced tasks like solving differential equations.

%%%%%%%%%%%%%%%%%%%%%%%%%%%%%%%%%%%%%%%%%%%%%%%%%%%%%%%%%%%%%%%%
\section{Related Work}
\textit{Linear Equations with Primitive Action Spaces}. Poesia et al.~\citep{poesia2021contrastive} pioneered the use of RL on symbolic equations. They showed agents trained with a contrastive loss can solve linear equations with integer coefficients. Their formulation differs from ours in that their action space is at a primitive granularity: states and actions are represented as raw text (strings of symbols), and each algebraic operation (associativity, commutativity, adding a constant, etc.) corresponds to a single low-level action. We represent equations as expression trees and define higher-level actions. They also use sparse rewards (we use dense rewards imbued with curiosity) and do not study nonlinear equations such as rational equations with complex functions (exp, sin, etc). See Appendix.

\textit{Q-Learning for Symbolic Equations}. Dabelow and Ueda \citep{dabelow2024symbolic} used Q-learning in a SymPy-based framework, restricting the agent to linear equation of form $a_0 + a_1 x = a_2 + a_3 x$ and $a_0 + b_0\, c \;+\; (a_1 + b_1\, c)\, x \;=\; a_2 + b_2\, c \;+\; (a_3 + b_3\, c)\ $ with coefficients in \(\mathbb{Z}, \mathbb{Q}, \mathbb{R}, \mathbb{C}\). They do not study nonlinear equations.

%%%%%%%%%%%%%%%%%%%%%%%%%%%%%%%%%%%%%%%%%%%%%%%%%%%%%%%%%%%%%%%%%%%%%%%%%%%%%%%%%%%%%%%%%%%%%%%%%%%%%%%%
\section{Problem Formulation}
Our Markov Decision Process (MDP) formulation is
\begin{wrapfigure}[12]{r}{0.55\columnwidth}
\vspace{-6pt}
\centering
\includegraphics[width=\linewidth]{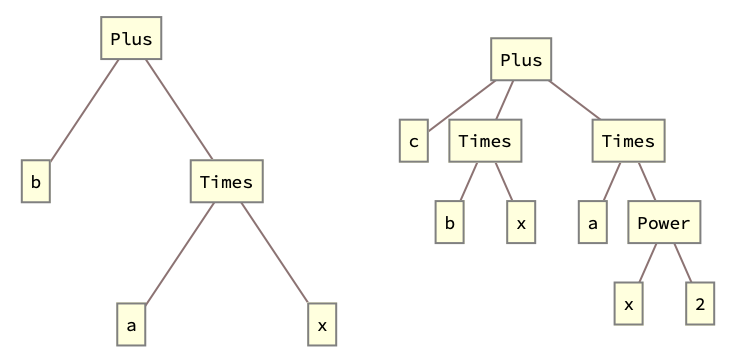} % fill the wrap box
\caption{Expression trees for \(ax+b\) and \(ax^2+bx+c\).}
\label{trees}
\vspace{-8pt}
\end{wrapfigure}

\textbf{States}: equations like $ax+b =0$ or $cx+d = -x/b$. We vectorize these using preorder traversal over the equations' expression tree (Figure~\ref{trees}) and pad to a maximum length $L=50$ (suitable length for the equation sizes we here consider). Then we map operations and symbols to integers \(\{\text{add}:1, \text{sub}:2, \text{mul}:3, \ldots, x:5, a:6, \ldots\}\). We encode both the lhs and rhs of the equation in this manner and then concatenate. An example embedding is $x+a \rightarrow [add, x, a] \rightarrow [1, 5, 8, PAD, \dots]$

\textbf{Actions}: represented as $(operation, term)$ pairs, such as $(sub, b)$ or $(div,a)$. Formally,
\begin{align}
O &= \{\,\op{add},\ \op{subtract},\ \op{mult},\ \op{divide}\,\} \cup  \{\op{square}, \ \op{sqrt},\ \op{exp},\ \op{log}, \op{sin}, \op{cos}\, \op{asin},\ \op{acos}\,\} \\
T   &= \SubExpr(\text{lhs}) \cup \SubExpr(\text{rhs}), \\
A   &= (O \times T) \cup \{(Expand, None),  (\op{collect}, x), (\op{multiply}, -1)\}.
\end{align}
Notice the first set of operations take two arguments, the second one argument. For the terms, we choose the list of sub-expressions in the lhs and rhs expression trees. Example of sub-expressions are
\begin{align}
& ax+b=0 \Rightarrow \{a, x, ax, b\} & \quad  \\
& (ax+b)/(cx+d)+e=0 \Rightarrow \{a, x, ax, b, c, d, cx, cx+d, ax+b\} 
\end{align}
This term set is expressive enough to solve rational equations and all other equation types we consider in this paper. It is also dynamic: the list of sub-expressions is derived from the equation/state and thus has variable length. Looping back to the operations, we also include $(mul,-1)$ and
\begin{align*}
    & \text{expand:} \quad cx + d + e(ax + b) \rightarrow aex + be + cx + d = 0 \\
    & \text{collect $x$:}  \quad  ax + bx \rightarrow (a + b)x \
    %& \text{multiply by $-1$}  
\end{align*}
These allow the agent to perform key algebraic steps (Appendix). Finally, we index the action set \(A\) serially, cap it at size \(|A|=50\) \footnote{We explored other values like $|A| = 40, 70, 100$ and found no changes in performance}, and mask illegal actions (e.g., division by zero, see Appendix). 

\textit{Rewards}. Define the \textit{complexity} $C$ of an equation as the total number of nodes and edges in the expression tree \footnote{We judged number of nodes+edges correlates better with algebraic complexity than number of nodes alone.}. The complexities of the equations in Figure~\ref{trees} are $C(ax+b) = 5 + 4 = 9$ and $C(ax^2 + bx + c) =  10 + 9 = 19$. The reward is then
\begin{align}
    R(\text{action}) = C(\text{equation}) - C(\text{equation after action})
\end{align}
The intuition here is to encourage the agent to take actions which simplify the equation. 

\textbf{State Transition Function}. We wrote our code in Python and used SymPy to apply operations to the equations. At each step, we keep track of a lhs and rhs of an equation e.g. \(lhs = (ax+b)\) and \(rhs = 0\). We apply actions to both the lhs and rhs. For example, $(sub,b)$ results in \((lhs, rhs) = (ax, -b)\), and then $(div,a)$ results in \((lhs, rhs) = (x, -b/a)\). The terminal condition for the environment is when \(lhs=x\) and the substitution of the $rhs$ into the original equation simplifies to $0$.

\textbf{Limitations}. Importantly, this MDP formulation only works on equations which are `closed,' in the sense that solving them requires manipulating the terms already present in the equation/sub-expression list. By contrast, solving 'open' equations requires adding new, out-of-equation terms or clever substitutions. A classic example is the quadratic equation $a x^2 + b x + c$ which is solved by completing the square -- adding $(b/2a)^2$ to each side. This is 'generative' reasoning, since the term $(b/2a)^2$ is \textit{not} in the term set we have defined. Equations that require these more exotic actions are beyond the scope of the current work (and were also not studied in all previous works \cite{poesia2021contrastive,dabelow2024symbolic}).

%%%%%%%%%%%%%%%%%%%%%%%%%%%%%%%%%%%%%%%%%%%%%%%%%%%%%%%%%%%%%%%%%%%%%%%%%%%%
\section{Results}
\textbf{Fixed equation environments}. We begin with six simple algebraic equations displayed in the first column of Table~\ref{Tsolve}. The first two, $ax+b, a/x+b$, are easily solved, and are included as baselines. The remaining four are more challenging, having a nested structure and more unknowns. Equations of this form are found in elementary textbooks on algebra, making them natural first test cases.
% \begin{table*}[t!]
% \centering
% \footnotesize % Further reduces font size
% \caption{Mean success rates ± standard errors over $N_{trial}=10$ runs and $ N_{train} = 3 \times 10^6$ steps.}
% \begin{tabular}{l c c c c c c c}
% \toprule
% Equation & A* & A2C & PPO & PPO-ICM & PPO-RIDE & PPO-NGU & PPO-RND \\
% \midrule
% $ax + b$ & $1.00 \pm 0.00$ & $1.00 \pm 0.00$ & $1.00 \pm 0.00$ & $1.00 \pm 0.00$ & $1.00 \pm 0.00$ & $1.00 \pm 0.00$ & $1.00 \pm 0.00$ \\
% $a / x + b$ & $1.00 \pm 0.00$ & $1.00 \pm 0.00$ & $1.00 \pm 0.00$ & $1.00 \pm 0.00$ & $1.00 \pm 0.00$ & $1.00 \pm 0.00$ & $1.00 \pm 0.00$ \\
% $c (a x + b) + d$ & \win{$1.00 \pm 0.00$} & $0.20 \pm 0.13$ & $0.10 \pm 0.09$ & $0.70 \pm 0.14$ & $0.60 \pm 0.15$ & $0.40 \pm 0.15$ & \win{$1.00 \pm 0.00$} \\
% $c + d / (a x + b)$ & $0.00 \pm 0.00$ & $0.00 \pm 0.00$ & $0.20 \pm 0.13$ & $0.50 \pm 0.16$ & $0.40 \pm 0.15$ & $0.50 \pm 0.16$ & \win{$0.80 \pm 0.13$} \\
% $(a x + b) + e (c x + d)$ & $0.00 \pm 0.00$ & $0.00 \pm 0.00$ & $0.00 \pm 0.00$ & $0.20 \pm 0.13$ & $0.00 \pm 0.00$ & $0.20 \pm 0.13$ & \win{$0.40 \pm 0.15$} \\
% $e + (a x + b) / (c x + d)$ & $0.00 \pm 0.00$ & $0.00 \pm 0.00$ & $0.00 \pm 0.00$ & $0.10 \pm 0.09$ & $0.00 \pm 0.00$ & $0.30 \pm 0.14$ & \win{$0.50 \pm 0.16$} \\
% \bottomrule
% \label{Tsolve}
% \end{tabular}
% \end{table*}

\begin{table*}[t]
\centering
%\scriptsize
\small
\caption{Mean success rates ± s.e. over $N_{\text{trial}}=10$, $N_{\text{train}}=3\times10^6$.}
\resizebox{\textwidth}{!}{%
\begin{tabular}{lccccccc}
\toprule
Equation & A* & A2C & PPO & PPO-ICM & PPO-RIDE & PPO-NGU & PPO-RND \\
\midrule
$ax + b$ & $1.00 \pm 0.00$ & $1.00 \pm 0.00$ & $1.00 \pm 0.00$ & $1.00 \pm 0.00$ & $1.00 \pm 0.00$ & $1.00 \pm 0.00$ & $1.00 \pm 0.00$ \\
$a / x + b$ & $1.00 \pm 0.00$ & $1.00 \pm 0.00$ & $1.00 \pm 0.00$ & $1.00 \pm 0.00$ & $1.00 \pm 0.00$ & $1.00 \pm 0.00$ & $1.00 \pm 0.00$ \\
$c (a x + b) + d$ & \win{$1.00 \pm 0.00$} & $0.20 \pm 0.13$ & $0.10 \pm 0.09$ & $0.70 \pm 0.14$ & $0.60 \pm 0.15$ & $0.40 \pm 0.15$ & \win{$1.00 \pm 0.00$} \\
$c + d / (a x + b)$ & $0.00 \pm 0.00$ & $0.00 \pm 0.00$ & $0.20 \pm 0.13$ & $0.50 \pm 0.16$ & $0.40 \pm 0.15$ & $0.50 \pm 0.16$ & \win{$0.80 \pm 0.13$} \\
$(a x + b) + e (c x + d)$ & $0.00 \pm 0.00$ & $0.00 \pm 0.00$ & $0.00 \pm 0.00$ & $0.20 \pm 0.13$ & $0.00 \pm 0.00$ & $0.20 \pm 0.13$ & \win{$0.40 \pm 0.15$} \\
$e + (a x + b) / (c x + d)$ & $0.00 \pm 0.00$ & $0.00 \pm 0.00$ & $0.00 \pm 0.00$ & $0.10 \pm 0.09$ & $0.00 \pm 0.00$ & $0.30 \pm 0.14$ & \win{$0.50 \pm 0.16$} \\
\bottomrule
\end{tabular}}
\label{Tsolve}
\end{table*}

We consider a \textit{fixed equation} environment, where the agent tried to solve a single equation during every episode. We used the A2C and PPO algorithms from Stable Baselines 3 \cite{raffin2021stable} and a homemade $A^*$ algorithm as a non-learning baseline (Appendix). We also attached four different curiosity methods to PPO:  Intrinsic Curiosity Module (ICM), Rewarding Impact-Driven Exploration (RIDE), Never Give Up (NGU), and Random Network Distillation (RND) \cite{pathak2017curiosity,burda2019exploration,badia2020never,raileanu2020ride,pathak2019disagreement}. These are implemented in the python package $rllte$ \cite{yuan2025rllte}. We hypothesize these exploration methods are needed to solve complex equations. Each algorithm was trained for $N_{train}=3 \times 10^6$ steps over $N_{trial}=10$ trials, and performance was evaluated based on success rates.

Table~\ref{Tsolve} shows all methods solved the first two equations, but performance diverged for the more complex equations. For $c + d/(ax+b)$, PPO-RND led with a 0.80 success rate, followed by PPO-ICM and PPO-NGU at 0.50, while $A^*$ failed. The most challenging equations, $(ax+b)+e(cx+d)$ and $e + (ax+b)/(cx+d)$, saw low success rates with PPO-RND achieving 0.40 and 0.50. Overall, PPO-NGU and PPO-RND were the top performers. The Appendix shows some example solution traces.

The takeaway from this problem is that our MDP formulation based on expression trees works and that curiosity-based exploration is required to solve non-elementary equations.

\textbf{Random equation environments}. Now we task the agent with solving a random equation during each episode. We generate train/test sets by starting with $x$ and applying actions recursively. Recall actions are $(operation, term)$ pairs. The operations are as before, but we restrict terms to $(a, b, c)$ (we excluded $d, e$ since we wanted to favor long, rather than wide equations). Applying a single action generated equations like $ax, \log(x), x+b$, applying 2 actions equations like $ax+b, \log(x)+d, (x+b)/c$ and so on. We threw out any equations that SymPy couldn't solve, and considered multi-root equations solved when just one root was found (e.g. $\sin(x) = 0$ is solved if the agent found $x=0$ rather than $x = 2 n \pi$). We created a small dataset which included all equations of $depth<4$ (size 3874) and sub-sampled to a $10^3/10^2$ train/test split, and a large dataset of all equations of $depth<5$ (size 15625) sub-sampled to a $10^4/10^3$ train/test split.
\begin{figure}
  \centering
  \includegraphics[width=\columnwidth]{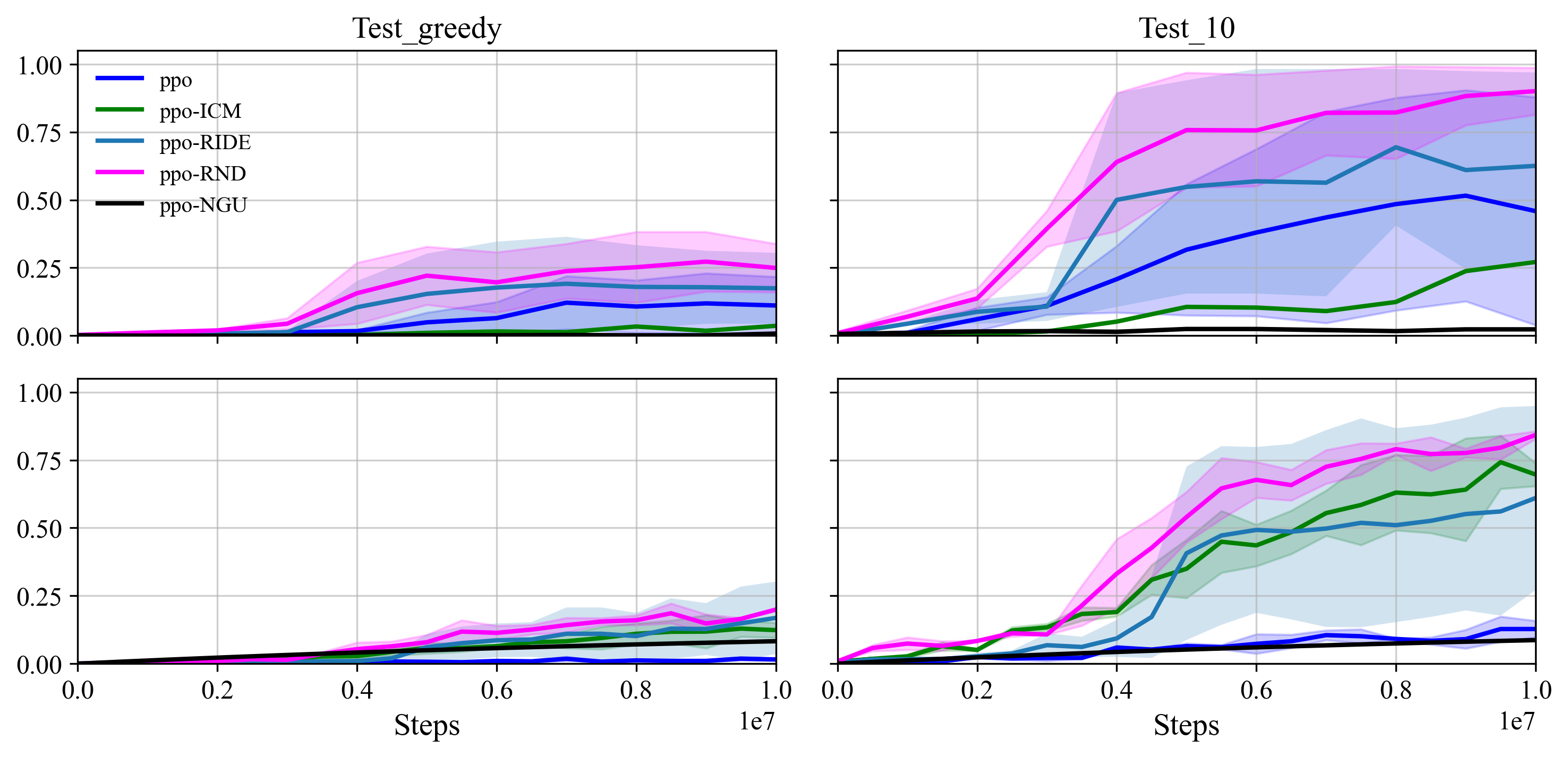}
  \caption{Top: small dataset; Bottom: large dataset. Curves show mean success rate over 3 trials; shading spans min–max range.}
  \label{learning_curves}
\end{figure}

Importantly, this generation process does \textit{not} create rational equations of form $ax+b/(cx+d)+e=0$. To include these, we augmented the dataset with a subset of such equations generated by combining simpler sub-expressions (Appendix). Below we give examples of equations and their solutions from the large dataset. %Overall, the dataset contains a diverse mix of elementary algebraic problems, making it a good test case for our purposes.
\begin{align*}
&a - b + \ln(x) = 0 \implies x = e^{b-a} && -b + \frac{-c + x/b}{c} = 0 \implies x = bc + bc^2 \\
&c + \sin\left(\frac{x-a}{c}\right) = 0 \implies x = a - c \arcsin(c) && \sin(\ln(\sin(x))) = 0 \implies x = \pi/2 \\
&\sqrt{bx+a}-c= 0 \implies x = (c^2-a) / b && b + \log(x^2/b^2) \implies x = \sqrt{b^2 \exp(-b)} \\
\end{align*}
Figure~\ref{learning_curves} plots the learning curves for both the small dataset (top row) and large dataset (bottom row). We plot $test_{greedy}$, the fraction of test equation solved following a greedy policy, and $test_{10}$ the fraction that were solved at least once from 10 rollouts following the stochastic policy. PPO-RND and PPO-RIDE are the best performers, achieving good generalization $test_{10} \approx 1.0, 0.8$ on both datasets (although $test_{greedy} \approx 0.25$ was much lower). PPO shows decent performance on the small datasets, but fails on the large one entirely. Surprisingly, PPO-NGU is the worst performer on this random equation environment, showing little progress on both the small and large datasets.

As before, the takeaway is that only curiosity-based exploration can solve the hardest tasks.

%%%%%%%%%%%%%%%%%%%%%%%%%%%%%%%%%%%%%%%%%%%%%
\section{Discussion}
\begin{wrapfigure}{r}{0.55\textwidth} % 'r' = right, 'l' = left
\vspace{-10pt} % adjust vertical spacing if needed
\begin{tcolorbox}[%
    title=Box 1: Open Challenges, 
    colback=yellow!30, 
    colframe=red, 
    sharpish corners, 
    width=\linewidth, 
    boxsep=1pt,
    coltitle=black,
    fonttitle=\bfseries,
    boxrule=0.5pt
]
\footnotesize
\begin{tabular}{@{}l c@{}}
\textbf{Algebraic equations} & \textbf{Solvable with RL} \\ \hline
\( a x^2 + b x + c = 0 \) & ? \\
\( a x^3 + b x^2 + c x + d = 0 \) & ? \\
\( a x^4 + b x^3 + c x^2 + d x + e = 0 \) & ?
\end{tabular}
\end{tcolorbox}
\vspace{-10pt}
\end{wrapfigure}
We have shown that curiosity-driven RL agents with action spaces based on expression trees can learn to solve some simple nonlinear equations in one variable, going beyond the linear equations considered in previous work \cite{poesia2021contrastive,dabelow2024symbolic}. Of the four curiosity types we tested, RND was the best performer. It is lightweight computationally, and solved both the fixed and random equation environments. NGU by contrast has a higher memory footprint, which may explain its poor performance on the random equation environment. Future work could explore this issue.

The main limitation of our work -- and all previous work -- is that the equations we considered were ``closed,''; all manipulations needed to solve them were contained within the original expression; the right $term$ in the action tuple $(operation, term)$ lay in the expression tree; we never had to generate novel terms. Extending our method to handle ``open'' equations, where new terms, auxiliary expressions, or changes of variables are introduced, will likely require a generative sub-model. 

To expand on this a little, Box 1 lists the first three non-trivial polynomial equations as benchmarks for future work. Admittedly, these are challenging problems. Solving the cubic and quartic equations requires an agent to discover Cardano's and Ferrari's method which are rather labyrinthine \cite{boyer2011history}. But solving the quadratic equation ought to be within reach. We speculate an Alphazero-like algorithm \cite{silver2017mastering}, which excels at long-horizon planning, might here be useful.

%%%%%%%%%%%%%%%%%%%%%%%%%%%%%%%%%%%%%%%%%%%%%%%%%%%%%%%%%%%%

%\section*{Impact Statement}
% This paper explores reinforcement learning for solving symbolic mathematical equations. Our goal is to advance methods for AI-driven mathematical reasoning. We do not foresee specific negative societal or ethical consequences arising directly from this work. Potential applications include education, scientific computing, and symbolic AI systems.
% This paper shows curiosity-driven reinforcement learning (RL) can solve some classes of nonlinear equations, advancing the development of AI-driven mathematical reasoning. Potential applications of the work are in education (e.g., intelligent tutoring systems that provide step-by-step equation-solving guidance), scientific computing (e.g., accelerating symbolic manipulation in computational modeling), and symbolic AI systems (e.g., supporting automated theorem proving or symbolic regression). We foresee no direct negative societal or ethical consequences from this research. 

\bibliography{ref}
\bibliographystyle{icml2025}

\newpage
\appendix
\onecolumn

\subsection{Macroactions}
Here we justify why the three macroactions $$(expand, none), (collect, x), (mul,-1)$$ discussed in the main text are \textit{required} to solve the equations we consider, and are not simple conveniences for the RL agent. Consider the action set \textit{without} these macroactions, which we requote below
\begin{align}
O &= \{\,\op{add},\ \op{subtract},\ \op{mul},\ \op{divide}\,\} \cup  \{\, \op{square}, \op{sqrt},\ \op{exp},\ \op{log}, \op{sin},\ \op{cos}\, \op{asin},\ \op{acos}\,\} \\
T   &= \SubExpr(\text{lhs}) \cup \SubExpr(\text{rhs}), \\
A   &= (O \times T) \cup \{(Expand, None)\} \{(\op{collect}, x)\} \cup \{(\op{multiply}, -1)\}.
\end{align}
Now consider the equation $dx + c(ax+b) = e$. One must distribute $c$ into the bracketed term $ax+b$. The action set above cannot do this. One needs the $expand$ macroaction for this. Moreover, once the term is expanded we get $dx+cax+cb = e$ and one needs to factor the $x$ common to both the first and second terms. $collect x$ performs this function. Finally, consider the equation $-x = b$. Since $-1$ is not included in our term set, one needs $(mul,-1)$ to solve the equation. (One could alternatively add $-1$ to the term set but we chose not to.)

\subsection{Dataset generation: Rational Equation  }

To supplement the recursive datasets, we constructed rational equations of the form
\[
\frac{a x + b}{c x + a} + b = 0,
\]
where \(a,b,c\) are (non-zero) symbolic coefficients. We excluded degenerate cases such as denominators that vanish identically or cancel trivially with numerators.

Examples of retained equations include:
\[
\frac{x+b}{c x + a} = 0 
\quad \Rightarrow \quad x = -b,
\]
\[
\frac{a x+b}{c x+a} + b = 0 
\quad \Rightarrow \quad x = \frac{-b - a b}{a + c b},
\]
\[
\frac{a x+b}{c x+a} - c = 0 
\quad \Rightarrow \quad x = \frac{-b + a c}{a - c^2}.
\]

These rational forms were merged into both the small and large datasets, with their proportion limited to preserve diversity across other functional forms (logarithmic, trigonometric, exponential, etc.).

\subsection{Illegal actions}
The main issue was to avoid illegal division by zero. This was not as simple as precluding $(div,0)$ from the action set. One had to prohibit `hidden' division by zero, such as prohibiting division by $(x+a)$ or $(x+b)$ for an equation of form $(x+a)(x+b)$. To do this, we checked if the equation of form $P(x)Q(x)..$, where $P(x),Q(x)$ are polynomials in $x$, and then removed $(div, P, Q, ...)$ from the action set.

\subsection{Hyperparameters}
We used Stable Baselines 3's default hyperparameters for A2C and PPO (Table~\ref{hyperparams}).

\begin{table}[h]
\small
\centering
\begin{tabular}{l p{0.7\textwidth}}
\toprule
\textbf{Algorithm} & \textbf{Hyperparameters} \\
\midrule
A2C & \(learning\_rate=0.0007\), \(n\_steps=5\), \(gamma=0.99\), \(gae\_lambda=1.0\), \(ent\_coef=0.0\), \(vf\_coef=0.5\), \(max\_grad\_norm=0.5\), \(use\_rms\_prop=True\), \(device='auto'\) \\
PPO & \(learning\_rate=0.0003\), \(n\_steps=2048\), \(batch\_size=64\), \(n\_epochs=10\), \(gamma=0.99\), \(gae\_lambda=0.95\), \(clip\_range=0.2\), \(ent\_coef=0.0\), \(vf\_coef=0.5\), \(max\_grad\_norm=0.5\), \(normalize\_advantage=True\), \(device='auto'\) \\
\bottomrule
\end{tabular}
\caption{Default hyperparameters for A2C and PPO from Stable Baselines 3.}
\label{hyperparams}
\end{table}

\subsection{A* Implementation}
A simple A* baseline searches the equation graph using complexity $C$ (nodes + edges) as heuristic. States are hashed via SymPy strings; priority queue expands lowest $g(n) + h(n)$ nodes. Depth limited to 20; actions masked for validity. Pseudocode: Initialize queue with start state; expand until solved or empty, updating scores and paths. Serves as non-learning baseline, efficient for shallow trees but limited by heuristic admissibility.

\subsection{Solution Traces}
Below are the solution traces for each equation in the fixed equation environment. Notice the solution trace is not always optimal. For instance, $ax+b=0$ could be solved with $(sub,b),(div,a)$, but instead the agent selects $(sub,ax)$ and then $(mul,-1)$. "Truediv" is the sympy notation for divide.
\begin{enumerate}
\item Equation: $a + x = 0$
\begin{verbatim}
Step 1: a + x = 0 | subtract, a
Solved: x = -a
\end{verbatim}
\item Equation: $a x = 0$
\begin{verbatim}
Step 1: a*x = 0 | div, a
Solved: x = 0
\end{verbatim}
\item Equation: $a x + b = 0$
\begin{verbatim}
Step 1: ax + b = 0 | subtract, ax
Step 2: -ax = b | multiply, -1
Step 3: ax = -b | div, a
Solved: x = -b/a
\end{verbatim}
\item Equation: $a/x + b = 0$
\begin{verbatim}
Step 1: a/x + b = 0 | subtract, b
Step 2: a/x = -b | truediv, 1/x
Step 3: -b*x = a | truediv, b
Step 4: -x = a/b | multiply, -1
Solved: x = -a/b
\end{verbatim}
\item Equation: $c(a x + b) + d = 0$
\begin{verbatim}
Step 1: c*(ax + b) + d = 0 | expand, None
Step 2: acx + bc + d = 0 | subtract, acx
Step 3: -acx = bc + d | multiply, -1
Step 4: acx = -bc - d | truediv, c
Step 5: ax = (-bc - d)/c | truediv, a
Solved: x = (-bc - d)/(ac)
\end{verbatim}
\item Equation: $c + d/(a x + b) = 0$
\begin{verbatim}
Step 1: c + d/(ax + b) = 0 | subtract, c
Step 2: d/(ax + b) = -c | multiply, (ax + b)
Step 3: d = -c (ax + b) | expand, None
Step 4: d = -c ax - c b | multiply, -1
Step 5: -d = c ax + c b | subtract, c b
Step 6: -d - c b = c ax | truediv, c
Step 7: (-d - c b)/c = ax | truediv, a
Step 8: ((-d - c b)/c)/a = x | truediv, 1/a
Solved: x = (-d - c b)/(c a)
\end{verbatim}
\item Equation: $c x + d + e(a x + b) = 0$
\begin{verbatim}
Step 1: cx + d + e(ax + b) = 0 | expand, None
Step 2: aex + be + cx + d = 0 | collect, x
Step 3: be + d + x*(ae + c) = 0 | subtract, x(ae + c)
Step 4: -x(ae + c) = be + d | truediv, ae + c
Step 5: -x = (be + d)/(ae + c) | multiply, -1
Solved: x = -(be + d)/(a*e + c)
\end{verbatim}
\item Equation: $e + (a x + b)/(c x + d) = 0$
\begin{verbatim}
Step 1: e + (ax + b)/(cx + d) = 0 | truediv, 1/(cx + d)
Step 2: (e + (ax + b)/(cx + d))(cx + d) = 0 | expand, None
Step 3: ax + b + cex + de = 0 | collect, x
Step 4: b + de + x*(a + ce) = 0 | subtract, x(a + ce)
Step 5: -x(a + ce) = b + de | expand, None
Step 6: -ax - cex = b + de | collect, x
Step 7: x*(-a - ce) = b + de | truediv, -a - ce
Solved: x = (b + de)/(-a - c*e)
\end{verbatim}
\end{enumerate}

\subsection{Comparison of our formulation to that of previous work by \cite{poesia2021contrastive}}

\begin{table}[h]
\centering
\small
\begin{tabular}{p{3.2cm}  p{5.1cm}  p{5.1cm}}
\toprule
\textbf{Aspect} &
\textbf{Poesia et al.\,(NeurIPS21)} &
\textbf{This work} \\
\midrule
Target domains &
Linear 1‑var equations (plus toy “CommonCore” envs) &
Linear \emph{and} nonlinear algebraic,\newline
trigonometric \\[2pt]

State representation &
Raw text $\rightarrow$ bi‑LSTM (character level) &
Expression–tree graph\newline
$\rightarrow$ MLP \\[2pt]

Action granularity &
Very primitive axioms (commute, distribute, add‑\(c\)…)\newline
$\Rightarrow$ huge branching factor &
Macro ops via \texttt{SymPy} (expand, collect x) $+$ arithmetic\newline
Dynamic term list = all sub‑expressions \\[2pt]

Reward signal &
\textbf{Sparse} — binary solved / unsolved &
\textbf{Dense} — drop in complexity $+\;$ curiosity \\[2pt]

Exploration / search &
Beam search during training;\newline
Contrastive Policy Learning (InfoNCE) on positives vs. beam negatives &
Pure model‑free A2C / PPO;\newline
(no external search) \\[2pt]

% Planner at inference &
% Greedy (or shallow search) &
% Greedy; no beam/heuristic needed \\[2pt]

Network backbone &
2‑layer bidirectional LSTM &
2‑layer MLP  \\[2pt] \\[2pt]

Core strength &
Bootstraps from \emph{extreme} reward sparsity with no shaping &
Scales to richer mathematics via structured representation and macro actions \\[2pt]
\bottomrule
\end{tabular}
\caption{Side‑by‑side comparison of our RL formulation with that of \citet{poesia2021contrastive}.}
\label{tab:poesia_comparison}
\end{table}

\end{document}